\documentclass[conference]{IEEEtran}
\IEEEoverridecommandlockouts

\usepackage{cite}
\usepackage{amsmath,amssymb,amsfonts}         
\usepackage{algorithm}
\usepackage{algpseudocode}
\usepackage{stfloats}
\usepackage{multirow}
\usepackage{bm}
\usepackage{array}
\usepackage{graphicx}
\usepackage{textcomp}
\usepackage{xcolor}
\usepackage{array}
\usepackage{pgfplots}
\usepackage{tikz}
\pgfplotsset{compat=newest}
\def\BibTeX{{\rm B\kern-.05em{\sc i\kern-.025em b}\kern-.08em
    T\kern-.1667em\lower.7ex\hbox{E}\kern-.125emX}}

\newcommand{\h}[1]{\textbf{#1}}

\usepackage[normalem]{ulem}

\begin{document}

\title{A Plug-and-Play Framework for Volumetric Light-Sheet Image Reconstruction}


\author{
\makebox[\linewidth][c]{
\begin{minipage}[t]{0.30\linewidth}\centering
1$^{st}$ Yi Gong\\
\textit{Department of Mathematics}\\
\textit{The University of North Carolina at Chapel Hill}\\
Chapel Hill, NC 27599, USA\\
\texttt{yigong@unc.edu}
\end{minipage}\hfill
\begin{minipage}[t]{0.31\linewidth}\centering
2$^{nd}$ Xinyuan Zhang\\
\textit{Department of Bioengineering}\\
\textit{The University of Texas at Dallas}\\
Richardson, TX 75080, USA\\
\texttt{xinyuan.zhang@utdallas.edu}
\end{minipage}\hfill
\begin{minipage}[t]{0.3\linewidth}\centering
3$^{rd}$ Jichen Chai\\
\textit{Department of Bioengineering}\\
\textit{The University of Texas at Dallas}\\
Richardson, TX 75080, USA\\
\texttt{jichen.chai@utdallas.edu}
\end{minipage}
}
\\[0.8em]
\makebox[\linewidth][c]{
\hspace{0.24\linewidth}
\begin{minipage}[t]{0.32\linewidth}\centering
4$^{th}$ Yichen Ding\\
\textit{Department of Bioengineering}\\
\textit{The University of Texas at Dallas}\\
Richardson, TX 75080, USA\\
\texttt{yichen.ding@utdallas.edu}
\end{minipage}\hfill
\begin{minipage}[t]{0.48\linewidth}\centering
5$^{th}$ Yifei Lou\\
\textit{Department of Mathematics}\\
\textit{The University of North Carolina at Chapel Hill}\\
Chapel Hill, NC 27599, USA\\
\texttt{yflou@unc.edu}
\end{minipage}
\hspace{0.15\linewidth}
}
}

\maketitle

\begin{abstract} Cardiac contraction is a rapid, coordinated process that unfolds across three-dimensional tissue on millisecond timescales. Traditional optical imaging is often inadequate for capturing dynamic cellular structure in the beating heart because of a fundamental trade-off between spatial and temporal resolution. To overcome these limitations, we propose a high-performance computational imaging framework that integrates Compressive Sensing (CS) with Light-Sheet Microscopy (LSM) for efficient, low-phototoxic cardiac imaging. The system performs compressed acquisition of fluorescence signals via random binary mask coding using a Digital Micromirror Device (DMD). We propose a Plug-and-Play (PnP) framework, solved using the alternating direction method of multipliers (ADMM), which flexibly incorporates advanced denoisers, including Tikhonov, Total Variation (TV), and BM3D. To preserve structural continuity in dynamic imaging, we further introduce temporal regularization enforcing smoothness between adjacent z-slices. Experimental results on zebrafish heart imaging under high compression ratios demonstrate that the proposed method successfully reconstructs cellular structures with excellent denoising performance and image clarity, validating the effectiveness and robustness of our algorithm in real-world high-speed, low-light biological imaging scenarios.

\end{abstract}

\begin{IEEEkeywords}
Compressive Sensing, Light-Sheet Microscopy, Plug-and-Play, Image Denoising, Temporal Regularization
\end{IEEEkeywords}

\section{Introduction}

Heart muscle cells, or cardiomyocytes, generate coordinated contractions that drive each heartbeat. Understanding how the heart develops, maintains rhythm, and responds to disease requires detailed imaging of contraction dynamics \textit{in vivo} \cite{bers2002cardiac,chen2019diamond,zhang2022computational}. Achieving this level of detail depends on specialized optical tools that can capture fast, 3D cardiac contraction with cellular resolution, while keeping light exposure low to avoid photo-damage. 

However, modern optical microscopy systems face significant challenges in achieving high-speed, high-resolution 3D reconstruction \textit{in vivo}, especially when constrained by the need to minimize phototoxicity and thermal load.
Current 3D imaging of contracting hearts primarily relies on confocal laser scanning microscopy (CLSM) \cite{Minsky1988AO,White1987JCB}, two-photon microscopy (TPM) \cite{Denk1990Science}, and light-sheet microscopy (LSM) \cite{Huisken2004Science,Santi2011JHC} together with its lattice variant \cite{Chen2014Science}. Specifically, CLSM uses point scanning and a pinhole to achieve high lateral resolution optical sectioning and has been widely applied in biological imaging. However, point-by-point acquisition limits volumetric speed, and the high excitation dose required to maintain the signal-to-noise ratio can induce photobleaching and phototoxicity during long-term or thick tissue imaging \cite{Icha2017Bioessays}. Similarly, TPM confines excitation to the focal plane via nonlinear two-photon absorption, enabling deep imaging in scattering tissue, but remains constrained by point/line scanning and can accumulate thermal and phototoxic effects at high repetition rates \cite{Helmchen2005NatMethods}. Lastly, LSM employs selective plane illumination with orthogonal detection to substantially reduce light dose and accelerate volumetric acquisition. However, without highly specialized hardware, even with parallel light-sheet strategies that illuminate multiple planes simultaneously, volumetric rates at or below 20 Hz \textit{in vivo} are typically achieved \cite{sacconi2022khz,dean2017imaging}. Recent efforts that combine LSM with compressive sensing (CS) have made progress but remain limited for fast \textit{in vivo} volumetrics. Spatially modulated LSM with compressive sensing reconstructs volumes from patterned illuminations yet requires multiple sequential exposures, resulting in volumetric rates below 1 Hz \cite{calisesi2019spatially}. Similarly, snapshot temporal compressive LSM recovers multiple frames at high temporal resolution from a single measurement but operates on a single light-sheet plane rather than full 3D stacks \cite{wang2023snapshot}. On the other hand, light field microscopy provides single-shot volumetric capture by multiplexing angular information onto one sensor, although the trade-off between angular and spatial sampling reduces effective spatial resolution and limits cellular resolution imaging \textit{in vivo} \cite{saberigarakani2025volumetric, Wang2021VCDLFM, Wang2021Hybrid}.

To capture cardiac contraction with higher spatiotemporal fidelity than existing 3D fluorescence microscopy techniques allow, our team has developed a light-sheet microscopy (LSM) platform optimized for fast, low-phototoxic volumetric acquisition \cite{zhang2025instantaneous}. This system extends our previous LSM design \cite{zhang2023cardiac,zhang2024_4d} by integrating the compressive sensing paradigm \cite{donoho2006compressed} to further improve spatiotemporal resolution and data efficiency. The system achieves multi-slice compressed imaging by synchronizing axial scanning and spatial light modulation with high temporal precision, allowing fluorescence signals from different depth planes to be encoded within a single exposure. These encoded signals are optically transmitted and captured at high frame rates, enabling volumetric imaging at up to 200 volumes per second while preserving cellular resolution. This hardware design overcomes the limitations of conventional light-sheet systems by enhancing imaging speed without compromising resolution or sample viability. Moreover, by operating at a given compression ratio, the required storage is reduced by the same factor, providing clear advantages for high-speed volumetric imaging.

Building on this fast data acquisition, we propose a reconstruction framework to address the challenges posed by nonlinear superimposition of depth layers, strong correlations across adjacent slices, and scanning-induced artifacts, which are the primary focus of this paper. Specifically, our imaging system is based on a compressed sensing acquisition strategy, where binary mask modulation on the digital micromirror device (DMD), which rapidly projects programmable binary patterns for spatial encoding, is combined with continuous axial scanning to encode fluorescence signals from multiple depth planes into each measurement. We enhance the resolution by exploiting spatial regularization and inter-slice correlations, which together define an underdetermined linear system arising from the compressive sensing strategy with DMD mask modulation and continuous axial scanning.

We employ a Plug-and-Play (PnP) framework with a generic regularization term to deal with the ill-posed inverse problem. The proposed model can be optimized efficiently using the alternating direction method of multipliers (ADMM) \cite{boyd2011distributed}, where one of the subproblems amounts to image denoising. The use of generic regularization provides the flexibility to incorporate various image priors from the literature, resulting in what we refer to as a slice-based model. We further impose temporal smoothness on the model. Computational efficiency is enhanced through a Woodbury-based inversion that accelerates large-scale matrix operations and a Gauss–Seidel sequential update for temporal regularization, which decouples slices while preserving structural continuity along the z-axis. Our experiments show that this temporal regularization consistently outperforms purely slice-based approaches.

\section{Literature review}
Regularization plays an indispensable role in solving inverse imaging problems such as image denoising. One classical approach is the Tikhonov regularization 
\cite{tikhonov1977solutions} that simply incorporates the $L_2$ norm squared penalty. Although effective for noise removal, this method often produces overly smooth results that blur important image features. 
To mitigate the smoothness, the total variation (TV) was proposed by Rudin-Osher-Fatemi \cite{rudin1992nonlinear} to pioneer variational denoising. The TV model formulates denoising as minimizing the $L_1$ norm of the image gradient under statistical noise constraints. Later, the TV minimization problem was accelerated by various techniques, including the Lagrangian method \cite{chambolle1997image}, a dual formulation \cite{chambolle2004algorithm}, and the split Bregman scheme  \cite{goldstein2009split}. Some TV variants include fractional-order total variation \cite{bai2007fractional,rahman2020poisson}, total generalized variation \cite{kongskov2019directional},  and a weighted difference model that combines anisotropic and isotropic TV \cite{lou2015weighted}.



To overcome the limitations of purely local regularization, such as the TV family, increasing attention has been directed toward incorporating sparse representations and non-local image priors. For example, dictionary learning methods aim to construct an overcomplete dictionary from image patches, using sparse coding and reconstruction to improve detail preservation. A representative approach is K-SVD \cite{elad2006image}, which alternates between updating the sparse coefficients and refining the dictionary atoms. This iterative process enables each patch to be accurately represented by a few basis vectors, effectively capturing local structural information. On the other hand,
Non-local Means (NLM) \cite{buades2005non-local} explicitly utilizes non-local self-similarity for image denoising  by computing a weighted average of all pixels whose surrounding patches are similar to that of the target, effectively overcoming the limitations of traditional local filters. However, because patch similarity is computed from noisy observations, the similarity weights can be severely affected by noise, leading to suboptimal averaging and loss of high-frequency details.
Alternatively, Block-matching and 3D filtering (BM3D)  \cite{dabov2007image} groups similar patches across the image into 3D stacks and applies collaborative filtering in the transform domain, which involves a 3D transformation, coefficient shrinkage, and inverse transformation. Unlike traditional methods that operate on individual patches, BM3D exploits group sparsity across collections of patches, resulting in improved noise suppression while preserving fine textures and structures. Its success demonstrates the complementarity of non-local self-similarity and sparse modeling.
Building on the non-local patch grouping strategy introduced in BM3D, Weighted Nuclear Norm Minimization (WNNM) \cite{gu2014weighted} introduces a more structured prior by representing non-local patch groups as matrices. It then applies weighted nuclear norm minimization to explicitly exploit their low-rank structure, thereby enhancing the preservation of textures and repeated patterns.

Recently, deep learning (DL) methods have achieved strong results in image denoising through end-to-end training. For instance, the Denoising Convolutional Neural Network (DnCNN) \cite{zhang2017beyond} leverages residual learning and deep CNN architectures, but such DL-based methods rely heavily on large amounts of training data. In contrast, the untrained Deep Image Prior (DIP) \cite{ulyanov2018deep} demonstrates that the structure of a randomly initialized convolutional network can itself capture natural image statistics during optimization, enabling denoising without any training data and inspiring unsupervised image restoration.

While advances in image denoising, ranging from variational methods to deep learning, have demonstrated impressive capabilities, many imaging tasks involve more general inverse problems beyond simple denoising. To address these challenges, the Plug-and-Play (PnP) framework \cite{venkatakrishnan2013plug} has emerged as a flexible approach, leveraging advanced denoisers as implicit priors within iterative optimization. In particular, PnP methods based on ADMM treat one of the subproblems as a denoising step, allowing the integration of a wide range of denoising algorithms while efficiently solving complex inverse problems. Chan et al. \cite{chan2017plug} established the fixed-point convergence theory of PnP-ADMM, proving that algorithms satisfying the conditions of a bounded denoiser converge to a fixed point under certain conditions, providing theoretical guarantees for the reliability of the method. The PnP framework has been successfully applied to various image restoration tasks, such as magnetic resonance imaging \cite{ahmad2020plug}, demonstrating its practical value and flexibility.

\section{Proposed Approaches}
After describing the image formation process in Section \ref{sect:forward}, we start with a slice-based image reconstruction model in Section \ref{sect:slice} that decomposes the reconstruction of a three-dimensional volumetric image into independent two-dimensional sub-problems. We then incorporate a smooth requirement along the $z$-direction, referred to as temporal correlation, into the proposed model, as detailed in Section \ref{sect:temporal}. For both approaches, we adopt a generic regularization term and solve the corresponding optimization problem via ADMM \cite{boyd2011distributed} due to its simplicity and efficiency.

\begin{figure*}
\centering
\includegraphics[width=1.25\columnwidth]{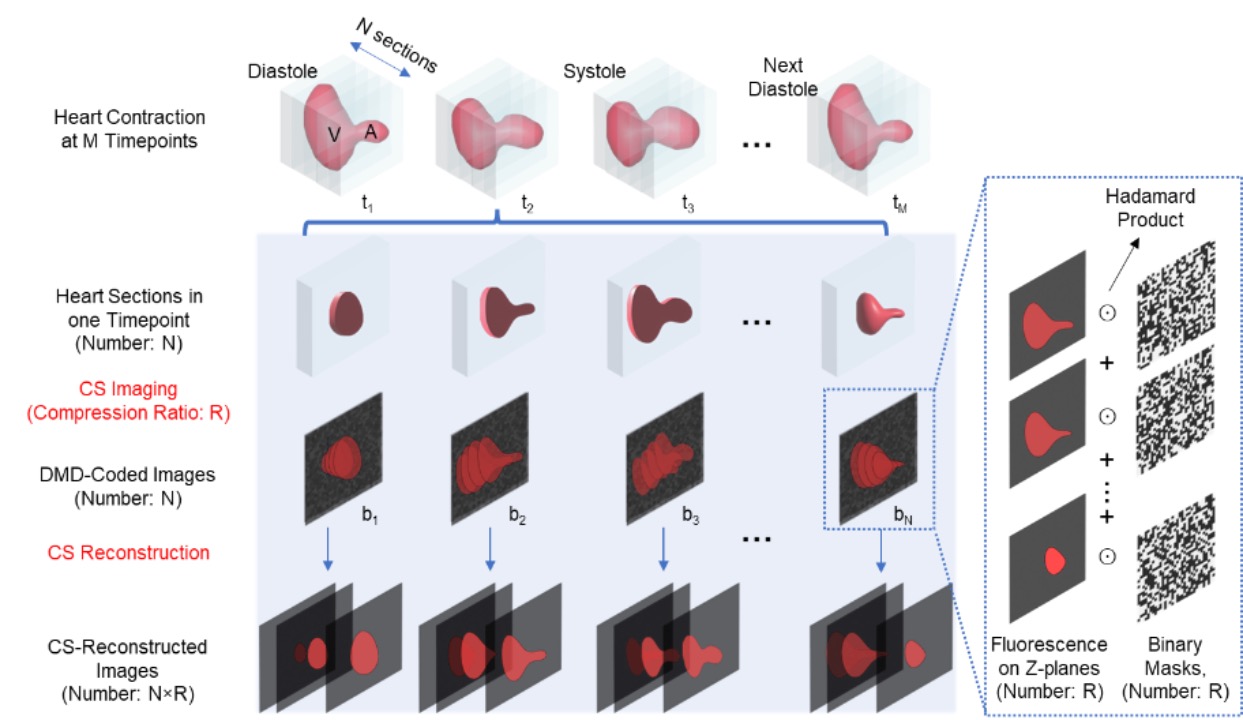}
\caption{Design of CS-LSM imaging formation.}
\label{fig:forward_model}
\end{figure*}
\subsection{Forward problem}\label{sect:forward}
We obtain compressed image data through a set of binary masks as shown in Fig. \ref{fig:forward_model}. For the beating zebrafish heart, we acquire 3D image volumes at $M$ distinct time points (e.g. $t_1, t_2, \dots, t_M$) throughout the cardiac cycle, such as during diastole and systole. Although the schematic depicts multiple time points throughout the cardiac cycle (e.g., $t_1, t_2, \dots, t_M$), we focus on one fixed time point for 3D reconstruction in this work. Note that the so-called ``temporal dimension'' refers to the stacking of slices along the axial $z$ direction, rather than different cardiac phases. Over a 10-ms interval, we acquire $N$ shots that together capture the full three-dimensional volume. Each shot records a subvolume formed by the sum of multiple encoded two-dimensional slices. Within a shot, each slice is modulated by a distinct binary mask displayed on the DMD, implemented as element-wise multiplication (Hadamard product) between the slice and the mask. The set of $N$ encoded shots is then used to reconstruct the complete volume. The number of slices combined into each coded image is determined by the compression ratio $R$.

In the compressed sensing process, the DMD applies $R$ binary masks sequentially during each camera exposure, optically encoding sequential $z$-planes with unique 2D patterns. These compressed images $\bm b_1, \bm b_2, \dots, \bm b_N$ are generated by summing multiple slices, where each image contains compressed information from multiple 2D slices.
The data $\bm b=[\bm b_1, \bm b_2, \cdots, \bm b_N]$ consist of these compressed images, which are obtained by recording partial sampling of the fluorescence data. The collected image data are sub-Nyquist sampled, meaning only a portion of the complete fluorescence data of the sample is recorded.

At the fixed time point, we present two reconstruction algorithms
to recover the complete 3D volume from these undersampled measurements. They utilized the prior information from the samples to recover the complete 3D cardiac volume, compensating for the artifacts and uncertainties introduced by the compressed data. The coded images are then processed by the reconstruction algorithm to recover the $N \times R$ images, which constitute the total volume of the heart.

\subsection{Slice-based Image Reconstruction}\label{sect:slice}

We consider a least-squares formulation for the data fidelity term and apply a generic regularization to each individual slice. Hence, the so-called ``slice-based'' image reconstruction can be formulated by
\begin{equation}\label{eq:model-slice}
\min_{\{\bm v_n\}}\sum_{j=1}^{N} \left\| \bm{b}_j - \sum_{r=1}^{R} \bm{\phi}_r \bm{v}_{(j-1)R + r} \right\|^2 + \lambda \sum_{n=1}^{NR} \psi(\bm{v}_n),
\end{equation}
where $\psi(\cdot)$ is a regularization term and $\lambda>0$ is a weighting parameter (to be tuned).  This model offers the flexibility to choose different regularization functions, allowing it to adapt to various image structures.

To minimize the proposed model \eqref{eq:model-slice}, we introduce a set of auxiliary variables $\bm u_n$ and consider an equivalent minimization problem as follows,
 \begin{equation}
    \begin{aligned}
    & \min_{\{\bm v_n,\bm u_n\}} \sum_{j=1}^{N} \left\| \bm{b}_j - \sum_{r=1}^{R} \bm{\phi}_r \bm{v}_{(j-1)R + r} \right\|^2 + \lambda \sum_{n=1}^{NR} \psi(\bm{u}_n) \\
 &\quad \mbox{subject to} \quad \bm{v}_n=\bm{u}_n. 
\end{aligned}
 \end{equation}
Its augmented Lagrangian is expressed by
\begin{multline}
 \mathcal{L}_{\rho}(\{\bm{v}_n\}, \{\bm{u}_n\}; \{\bm{d}_n\}):=\sum_{j=1}^{N} \left\| \bm{b}_j - \sum_{r=1}^{R} \bm{\phi}_r \bm{v}_{(j-1)R + r} \right\|^2 \\
+  \sum_{n=1}^{NR} \left(\lambda\psi(\bm{u}_n) + \frac{\rho}{2} \|\bm{v}_n-\bm{u}_n+\bm{d}_n\|^2\right),   
\end{multline}
where $\{\bm{d}_n\}$ is a set of dual variables and $\rho>0$ is a tunable parameter. The ADMM scheme iterates as follows, 
\begin{equation}\label{eq:ADMM}
\left\{\begin{array}{l}
     \{\bm{v}_n^{k+1}\} = \arg\min_{\{\bm{v}_n\}} \mathcal{L}_{\rho}(\{\bm{v}_n\}, \{\bm{u}_n^k\}; \{\bm{d}_n^k\}) \\
    \{\bm{u}_n^{k+1}\}= \arg\min_{\{\bm{u}_n\}} \mathcal{L}_{\rho}(\{\bm{v}_n^{k+1}\}, \{\bm{u}_n\}; \{\bm{d}_n^k\}) \\
     \bm{d}_n^{k+1} = \bm{d}_n^k + \bm{v}_n^{k+1} - \bm{u}_n^{k+1},
\end{array}
\right.
\end{equation}
where the superscript $k$ counts the iteration number.

For the $\bm{v}$-subproblem in \eqref{eq:ADMM}, we reconstruct a set of images corresponding to the data $\bm{b}_j$ simultaneously, i.e.,
\begin{multline}\label{eq:v-sub-1}
 \min_{\bm v}   \left\| \bm{b}_j - \sum_{r=1}^{R} \bm{\phi}_r \bm{v}_{(j-1)R + r} \right\|^2 
    + \frac {\rho} 2\sum_{r=1}^R \|\bm{v}_{(j-1)R + r}-\bm{g}_r\|^2,
\end{multline}
where we stack $  \{\bm{v}_{(j-1)R + r}^{k+1}\}$ into a long vector $\bm{v}$ and set $\bm{g}_r = \bm{u}^{k}_{(j-1)R + r}-\bm{d}^k_{(j-1)R + r}.$ Note that \eqref{eq:v-sub-1} has a closed-form solution, 
\begin{equation}\label{eq:v-update}
\bm{v}=(\Phi^T\Phi + \rho I)^{-1} (\Phi^T \bm{b}_j+ \rho \bm{g}),
\end{equation}
where $\Phi =[\phi_1, \cdots, \phi_r]$ and $\bm{g}=[\bm{g}_1, \cdots, \bm{g}_r].$ 
The inversion in \eqref{eq:v-update} can be simplified by applying the Woodbury matrix identity:
\begin{equation}\label{Woodbury}
(I + UV)^{-1} = I - U(I + VU)^{-1}V.
\end{equation}
Specifically, we can rewrite the matrix to be inverted by
\begin{equation}
(\Phi^T \Phi + \rho I)^{-1} = \frac{1}{\rho}\left(I + \frac{1}{\rho}\Phi^T \Phi\right)^{-1}.
\end{equation}
Applying \eqref{Woodbury}  with $U = \frac{1}{\rho}\Phi^T$ and $V = \Phi$ yields
\begin{equation}
\left(I + \frac{1}{\rho}\Phi^T \Phi\right)^{-1} = I - \frac{1}{\rho}\Phi^T\left(I + \frac{1}{\rho}\Phi\Phi^T\right)^{-1}\Phi,
\end{equation}
which can be expressed as,
\begin{equation}
(\Phi^T \Phi + \rho I)^{-1} = \frac{1}{\rho}\left[I - {\Phi^T\left(\rho I + \Phi\Phi^T\right)^{-1}}\Phi\right].
\end{equation}

Since each $\phi_r$ for $r=1, \cdots, R$ is diagonal, then $\Phi\Phi^T+\rho I$ is a diagonal matrix and hence its inversion can be calculated element-wise. Therefore, \eqref{eq:v-update} can be rewritten as
\begin{align*}
\bm{v} = &\frac{1}{\rho}\left[I - \Phi^T \left(\rho I + \Phi\Phi^T\right)^{-1} \Phi\right] \Phi^T \bm{b} \\
&\qquad+ \left[I - \Phi^T \left(\rho I + \Phi\Phi^T\right)^{-1} \Phi\right] \bm{g}\\
= &\frac{1}{\rho}\Phi^T \bm{b} - \frac{1}{\rho}\Phi^T \left(\rho I + \Phi\Phi^T\right)^{-1} \left(\Phi\Phi^T\right) \bm{b} \\
&\qquad + \left[I - \Phi^T \left(\rho I + \Phi\Phi^T\right)^{-1} \Phi\right] \bm{g}\\
= &\frac{1}{\rho}\Phi^T \bm{b} \left(1 - \frac{\left(\Phi\Phi^T\right)}{\rho + \left(\Phi\Phi^T\right)}\right) + \bm{g} - \Phi^T \left(\frac{\Phi\bm{g}}{\rho + \left(\Phi\Phi^T\right)}\right),\\
\end{align*}
thus leading to the update
\begin{equation}
  \bm{v}  = \frac{\Phi^T \bm{b}}{\rho + \Phi\Phi^T} + \bm{g} - \Phi^T \left(\frac{\Phi\bm{g}}{\rho + \Phi\Phi^T}\right).
\end{equation}

The $\bm{u}$-subproblem in \eqref{eq:ADMM} can be solved independently for each slice indexed by $n$, i.e.,
\begin{equation}\label{eq:u-sub}
\min_{\bm u_n} \lambda\psi(\bm{u}_n) + \frac{\rho} 2 \|\bm{v}_n-\bm{u}_n+\bm{d}_n\|^2,
\end{equation}
which can be regarded as a denoising step. In other words, the update of $\bm{u}_n$ amounts to denoising $\bm{v}_n+\bm{d}_n$ by a specific regularization, such as Tikhonov regularization, TV \cite{chambolle2004algorithm,goldstein2009split}, BM3D \cite{dabov2007image}, and WNNM 
\cite{gu2014weighted}. Specifically, for Tikhonov regularization, $\psi(\bm{u}) = \|\bm{u}\|^2$, the $\bm{u}$-subproblem in \eqref{eq:u-sub} is equivalent to 
\begin{equation}
\bm u_n = \arg\min_{\bm u} \frac{2\lambda + \rho}{2}\left\|\bm{u} - \frac{\rho(\bm{v}_n + \bm{d}_n)}{2\lambda + \rho}\right\|^2,
\end{equation}
which has a closed-form solution given by 
\[
\bm u_n^{\mbox{Tik}} = \frac{\rho(\bm{v}_n + \bm{d}_n)}{2\lambda + \rho}.
\]
Thanks to this closed-form solution, Tikhonov is more efficient than other denoising methods.

\subsection{Reconstruction with Spatial-Temporal Correlation}\label{sect:temporal}
We propose to further incorporate the temporal smoothing by minimizing the difference between two adjacent slices into the slice-based model \eqref{eq:model-slice}, thus leading to the following objective function,
\begin{multline}
\frac{1}{2} \sum_{j=1}^{N} \left\| \bm{b}_j - \sum_{r=1}^{R} \bm{\phi}_r \bm{v}_{(j-1)R + r} \right\|^2 
+ \lambda \sum_{n=1}^{NR} \psi(\bm{v}_n) \\
+ \frac{\gamma}{2} \sum_{n=1}^{NR+1} \left\| \bm{v}_n - \bm{v}_{n-1} \right\|^2,
\end{multline}
where $\gamma>0$ is a tunable parameter and we assume a periodic condition in the sense that $\bm{u}_0 = \bm{u}_{NR}$ and $\bm{u}_1= \bm{u}_{NR+1}$. This joint spatial-temporal constraint better preserves the continuity and consistency of image sequences, making it particularly suitable for dynamic imaging scenarios.

We consider an equivalent form 
\begin{multline}
\frac{1}{2}\sum_{j=1}^{N}\left\|\bm{b}_j - \sum_{r=1}^{R}\bm{\phi}_r \bm{v}_{(j-1)R+r}\right\|^2 + \lambda\sum_{n=1}^{NR}\psi(\bm{u}_n) \\
+ \frac{\gamma}{2}\sum_{n=1}^{NR+1}\|\bm{u}_n - \bm{u}_{n-1}\|^2 \quad \text{subject to } \bm{v}_n = \bm{u}_n, 
\end{multline}
and construct its augmented Lagrangian as
\begin{multline}
\mathcal{L}_\rho(\{\bm{v}_n\}, \{\bm{u}_n\}, \{\bm{d}_n\}) := \frac{1}{2}\sum_{j=1}^{N}\left\|\bm{b}_j - \sum_{r=1}^{R}\bm{\phi}_r \bm{v}_{(j-1)R+r}\right\|^2 \\
+ \sum_{n=1}^{NR}\left(\lambda\psi(\bm{u}_n) + \frac{\rho}{2}\|\bm{v}_n -\bm{u}_n + \bm{d}_n\|^2\right)  \\
+ \frac{\gamma}{2}\sum_{n=1}^{NR+1}\|\bm{u}_n - \bm{u}_{n-1}\|^2 ,
\end{multline}
where $\{\bm{d}_n\}$ is a set of dual variables and $\rho > 0$ is a parameter.  The ADMM iterations are the same as \eqref{eq:ADMM} with the same update rule for the $\bm v$-subproblem, i.e., \eqref{eq:v-update}.

Due to the coupled temporal term, the $\bm u$-subproblem can not be solved independently. Rather, we adopt the Gauss–Seidel scheme to update the $n$th slice while fixing its adjacent slices ($n-1$ and $n+1$). In particular, the minimization problem for the $n$th slice can be expressed by
\begin{align}
  \bm u_n=&\arg\min_{\bm u}\lambda\psi(\bm{u}) + \frac{\rho}{2}\|\bm{u}-(\bm{v}_n   + \bm{d}_n)\|^2\notag\\
  &\qquad \qquad 
 + \frac{\gamma}{2}\|\bm{u} -\bm{u}_{n-1}\|^2 + \frac{\gamma}{2}\|\bm{u}-\bm{u}_{n+1}\|^2. \label{eq:temporal_u_sub}
\end{align}
By completing the squares, the problem \eqref{eq:temporal_u_sub} is equivalent to
\begin{equation}\label{eq:u-update}
    \lambda \psi(\bm{u}) + \frac{\rho + 2\gamma}{2} \left\| 
\bm{u} - \frac{ \rho (\bm{v}_n + \bm{d}_n) + \gamma (\bm{u}_{n-1} +\bm{u}_{n+1}) }{ \rho + 2\gamma }
\right\|^2,
\end{equation}
which can be regarded as denoising the data $\frac{1}{\rho + 2\gamma} (\rho (\bm{v}_n + \bm{d}_n) + \gamma (\bm{u}_{n-1} +\bm{u}_{n+1}))$ by using the regularization $\psi.$ Take the Tikhonov regularization for an example. The $\bm{u}$-subproblem can be rewritten by
\[
\frac{2\lambda + \rho + 2\gamma}{2}\left\|\bm{u} - \frac{\rho(\bm{v}_n + \bm{d}_n) + \gamma(\bm{u}_{n-1} + \bm{u}_{n+1})}{2\lambda + \rho + 2\gamma}\right\|^2,
\]
thus leading to a closed-form solution
\[
\bm u_n^{\mbox{Tik}} = \frac{\rho(\bm{v}_n + \bm{d}_n) + \gamma(\bm{u}_{n-1} + \bm{u}_{n+1})}{2\lambda + \rho + 2\gamma}.
\]
We summarize the pseudo-code of the proposed temporal CS LSM reconstruction method in Algorithm \ref{alg:cs-lsm-pnp}. If we set the parameter $\gamma=0,$ it reduces to the slice-based approach. 

\begin{algorithm}[H]
\caption{CS LSM Temporal  Reconstruction Algorithm.}
\label{alg:cs-lsm-pnp}
\begin{algorithmic}
\Require Data $\{\h b_j\}$ for $j=1,\cdots, N$ and binary masks $\{\phi_r\}$ for $r=1, \cdots, R$. 

\State Set parameters $\lambda, \rho, \gamma \in \mathbb{R}^+$.

\State Initialize $\h v_n^0, \h u_n^0, \h d_n^0$ for $n=1,\cdots, N,$ and $k=0.$

\While{stopping conditions do not satisfy}
\For{$j = 1$ to $N$}
\State Collect the slices from $n=(j-1)R+1$ to $jR$
  \State Update $\h v_n^{k+1}$  all at   once via  \eqref{eq:v-update}.
\EndFor


  \For{$n = 1$ to $NR$}
  \State Let $\h g=\frac{1}{\rho + 2\gamma} (\rho (\bm{v}_n^{k+1} + \bm{d}_n^{k}) + \gamma (\bm{u}_{n-1}^{k+1} +\bm{u}_{n+1}^k))$
\State $\h u_n^{k+1}$ is a denoised version of $\h g$ via \eqref{eq:u-update}.
  \EndFor

  \For{$n = 1$ to $NR$}
  \State $\h d_n^{k+1} = \h d_n^k + \h v_n^{k+1} -\h u_n^{k+1}$
  \EndFor

  \State $k$ $\gets$  $k+ 1$
\EndWhile

\State \Return $\{\h v_n^k\}$ for $n=1, \cdots, NR.$
\State \textbf{Note:} When $\gamma = 0$, the algorithm reduces to slice-based processing without temporal coupling.
\end{algorithmic}
\end{algorithm}


\section{Experiments}
To assess reconstruction performance, we adopt two standard quantitative metrics: Peak Signal-to-Noise Ratio (PSNR) \cite{hore2010psnr} and Structural Similarity Index (SSIM) \cite{wang2004ssim}.

The PSNR evaluates the fidelity between the original image $I\in\mathbb R^{m\times n}$ and its reconstruction $\hat I$ by measuring the pixel-wise discrepancy, which is defined by
\begin{equation}
    \text{PSNR}(\hat{I}, I) = 10 \log_{10} \left( \frac{mn V^2}{\|\hat{I} - I\|_2^2} \right),
\end{equation}
where $V$ is the maximum possible intensity value in the original image and $m \times n$ denotes the image dimension. A higher PSNR suggests that the reconstruction more accurately preserves the pixel-level details of the original.

SSIM is used to assess perceptual quality by modeling the structural similarity between images in a way that aligns with the human visual system (HVS). Given a pair of corresponding local patches $p$ and $\hat{p}$, extracted from the original image $I$ and the reconstructed image $\hat{I}$, the local SSIM is defined by
\begin{equation}
    \text{ssim}(p, \hat{p}) = \frac{(2\mu_p \mu_{\hat{p}} + c_1)(2\sigma_{p\hat{p}} + c_2)}{(\mu_p^2 + \mu_{\hat{p}}^2 + c_1)(\sigma_p^2 + \sigma_{\hat{p}}^2 + c_2)},
\end{equation}
where $\mu_p$ and $\mu_{\hat{p}}$ are the mean intensities, $\sigma_p^2$ and $\sigma_{\hat{p}}^2$ the variances, and $\sigma_{p\hat{p}}$ the covariance between the two patches. The constants $c_1$ and $c_2$ are used to avoid instability in low contrast regions. All these parameters ($\mu_p, \mu_{\hat{p}}, \sigma_p^2, \sigma_{\hat{p}}^2,\sigma_{p\hat{p}}, c_1, c_2$) are user-specific. In our experiments, we choose $\mu_p$, $\mu_{\hat{p}}$, $\sigma_p^2$, $\sigma_{\hat{p}}^2$, and $\sigma_{p\hat{p}}$ computed over a $11 \times 11$ Gaussian window with standard deviation $\sigma = 1.5$, and constants $c_1 = (0.01 \cdot L)^2$, $c_2 = (0.03 \cdot L)^2$, where $L = 1$ for normalized input images. Then the general SSIM between two images is obtained by averaging the local SSIM values over all $L$ patch pairs:
\begin{equation}
    \text{SSIM}(\hat{I}, I) = \frac{1}{L} \sum_{\ell=1}^{L} \text{ssim}(p_\ell, \hat{p}_\ell),
\end{equation}
where $p_\ell$ and $\hat{p}_\ell$ denote the corresponding patches at location $\ell$ in the original and reconstructed images, respectively.
While PSNR quantifies absolute pixel-wise error, SSIM offers a perceptually aligned evaluation by capturing structural attributes such as edges and textures. An SSIM value close to 1 indicates that the reconstructed image retains high structural similarity to the original.

In our experiments, both ground truth and reconstructed images are 3D volumes. To evaluate reconstruction quality, we extend PSNR and SSIM from 2D to 3D. Specifically, PSNR is computed by treating all voxels in the 3D volume as a single array and measuring the overall pixel-wise discrepancy between the ground truth and reconstructed volumes, following the same formulation as in the 2D case. SSIM is extended by applying a $11 \times 11 \times 11$ Gaussian-weighted window (with standard deviation $\sigma = 1.5$) across the 3D space. At each voxel location, local statistics are computed over its neighborhood to obtain a local SSIM score, and the final SSIM value is obtained by averaging these scores over the entire volume.

\subsection{Data Generation}
To support a rigorous evaluation of our reconstruction algorithms, we synthesize a 3D dataset that reproduces zebrafish cardiac dynamics over a full heartbeat based on light-sheet imaging \cite{zhang2023cardiac}. This dataset is designed to provide precise ground truth for voxel-wise benchmarking, with explicit modeling of each stage in the simulation pipeline, including geometry, motion, optics, and simulated photon noise. For ground truth, we focus on a representative single time point defined as a 200×200×150-voxel volume containing both atrium and ventricle, with approximately 300 nuclei distributed adjacent to the chamber surfaces. In experiments, a total of 40 frames is selected to represent the volume. Experiments are performed under both noise-free and noisy conditions. In the noisy case, the simulated noise is assumed to be zero-mean Gaussian with variance 0.001, added to the compressed frames after intensity normalization and rescaling according to the compression ratio.

\subsection{Parameter Tuning}

To systematically evaluate the reconstruction performance under both noise-free and noisy conditions, we compare Tikhonov, TV, and BM3D priors within slice-based and temporal reconstruction frameworks. All competing methods are subject to identical stopping criteria to ensure fair comparison. For the noise-free scenario, we impose stringent convergence requirements to achieve high-precision solutions: the maximum number of iterations is capped at 100, with a relative convergence tolerance of 0.001. In contrast, for noisy data, the algorithm employs moderately relaxed convergence thresholds to balance reconstruction fidelity with numerical stability, allowing up to 200 iterations and setting the relative tolerance to 0.01.

To find the best combinations of parameters for each model, we employ a Bayesian optimization technique \cite{snoek2012practical} that takes advantage of user-defined search ranges and a performance-based objective function to identify optimal parameter configurations. The method employs a probabilistic surrogate model to approximate the objective landscape and iteratively selects promising candidates using an enhanced acquisition function, Expected Improvement Plus (EI+). EI+ extends the classical Expected Improvement (EI) \cite{jones1998efficient} by incorporating an anti-exploitation mechanism to improve sampling efficiency and reduce the likelihood of premature convergence to local optima. Specifically, it introduces a penalization term to discourage redundant evaluations and a diversity-aware component that promotes broader exploration of the search space. Compared to traditional grid search, Bayesian optimization leverages historical results to construct surrogate models, achieving a superior balance between exploration and exploitation while reducing computational overhead. 


We configure Bayesian optimization to perform a maximum of 50 function evaluations, with the objective defined as the minimization of negative PSNR to maximize reconstruction quality. Upon completion of the automated parameter tuning process, the optimal parameter configuration is extracted and used to re-run the algorithm, yielding the final reconstruction results.

\subsection{Noise-free Results}
We begin by evaluating the reconstruction performance on noise-free data, focusing on the optimal parameters, quantitative results, and visual comparisons for a representative frame.

\begin{table}[H]
\centering
\caption{Optimal parameters in the noise-free case.}
\begin{tabular}{|c|
>{\centering\arraybackslash}p{1.4cm}|
>{\centering\arraybackslash}p{1.4cm}|
>{\centering\arraybackslash}p{1.4cm}|}
\hline
\textbf{Parameter} & \textbf{$\lambda$} & \textbf{$\rho$} & \textbf{$\gamma$} \\
\hline
\multicolumn{4}{|c|}{\textbf{Slice-Based}} \\
\hline
Tikhonov & 0.0152 &  0.0010  & - \\
\hline
TV & 0.0174 & 0.0010 & - \\
\hline
BM3D & 9.6690 & 0.0031 & - \\
\hline
\multicolumn{4}{|c|}{\textbf{Temporal}} \\
\hline
Tikhonov & 0.0070 & 0.1000 & 0.5959 \\
\hline
TV & 0.0987 & 0.0114 & 0.0010 \\
\hline
BM3D & 9.4132 & 0.0960 & 0.0010 \\
\hline
\end{tabular}
\label{tab:NF-Optimal Parameter}
\end{table}

The optimal parameters obtained for each method are summarized in Table \ref{tab:NF-Optimal Parameter}, which shows that the choice of $\lambda$ for Tikhonov and TV is on the order of $10^{-2}$. This indicates that it is reasonable not to enforce strong regularization in the noise-free case. Note that the BM3D implementation \cite{dabov2007image} we adopt uses a different setting compared to \eqref{eq:u-sub}, in the sense that $\lambda$ is not a weighting parameter as in TV, but rather an indicator of the estimated noise level. Moreover, the temporal variants generally adopt stronger regularization than their slice-based counterparts, indicating that the incorporation of temporal smoothness necessitates larger penalty weights to stabilize the reconstruction.

\begin{table}[H]
\centering
\caption{Quantitative comparison of reconstruction from the noise-free data. The best results are shown in bold.}
\begin{tabular}{|c|c|c|c|c|}
\hline
 & \multicolumn{2}{c|}{PSNR (dB)} & \multicolumn{2}{c|}{SSIM} \\
\cline{2-5}
 & Slice-Base & Temporal & Slice-Base & Temporal \\
\hline
Tikhonov & 31.2328 & 34.8943 & 0.7470 & 0.9442 \\
\hline
TV & 38.1663 & 38.2551 & 0.9805 & 0.9794 \\
\hline
BM3D & 42.4688 & \textbf{42.6759} & 0.9918 & \textbf{0.9923} \\
\hline
\end{tabular}
\label{tab:results_without_noise}
\end{table}

\begin{figure*}[t]
    \centering
    \begin{tabular}{@{}c@{\hspace{25pt}}c@{\hspace{25pt}}c@{}}
        Ground Truth & Slice-base TV & Slice-base BM3D \\[9pt]
        \includegraphics[width=0.22\textwidth]{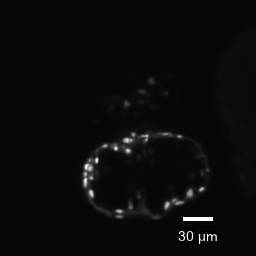} &
        \includegraphics[width=0.22\textwidth]{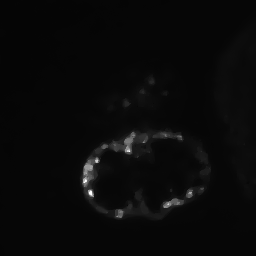} &
        \includegraphics[width=0.22\textwidth]{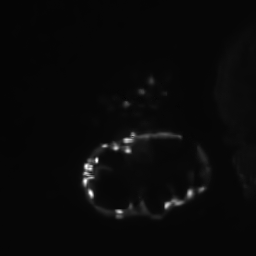} \\[9pt]   
        Temporal Tikhonov & Temporal TV & Temporal BM3D \\[9pt]
        \includegraphics[width=0.22\textwidth]{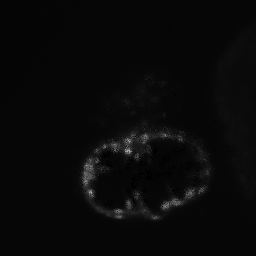} &
        \includegraphics[width=0.22\textwidth]{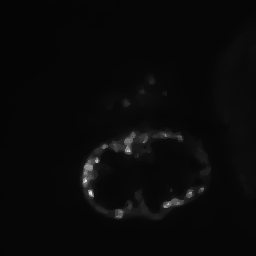} &
        \includegraphics[width=0.22\textwidth]{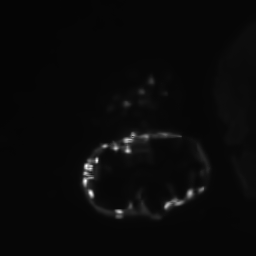} \\
    \end{tabular}
    \caption{Image reconstruction from noise-free data, the 17th slice is plotted.}
    \label{fig:NoNoiseResultsImage}
\end{figure*}

With the optimal parameters, we report the corresponding PSNR and SSIM values in Table \ref{tab:results_without_noise}. The results show that BM3D consistently achieves the best reconstruction performance, with Temporal-BM3D reaching a PSNR of 42.6759 dB and an SSIM of 0.9923, clearly outperforming all other methods. TV produces intermediate performance, maintaining satisfactory SSIM values (around 0.98) but noticeably lower PSNR than BM3D. Tikhonov performs the worst in both metrics. Across all priors, the temporal models outperform their slice-based counterparts, confirming that temporal correlation effectively enhances both stability and fidelity in the reconstruction.

Beyond the quantitative results, we select the 17th frame, which contains rich structural information, as a representative example to qualitatively investigate the visual results. Each white dot is an individual nucleus of the cardiomyocyte in the images. As shown in Fig.~\ref{fig:NoNoiseResultsImage}, BM3D reconstructions, particularly in the temporal setting, exhibit the clearest boundaries and faithfully preserve fine structures among all methods, while simultaneously suppressing background artifacts. In contrast, TV produces smoother outputs with relatively preserved global structures, while it suffers from noticeable edge blurring and partial loss of fine details. Temporal Tikhonov instead shows the notably weakest visual fidelity, characterized by low contrast and insufficient recovery of sharp features, which is consistent with its poor quantitative scores. Across the reconstructed images of all priors, the temporal variants produce visually sharper and more stable reconstructions than their slice-based counterparts, with the advantage being most prominent for BM3D, where temporal correlation further enhances edge clarity and structural consistency.

\subsection{Noisy Data Reconstruction}
We then examine the robustness of different priors when the measurements are corrupted. Compared with the noise-free case (Table \ref{tab:results_without_noise}), the overall reconstruction quality reported in Table \ref{tab:results_with_noise} deteriorates across all methods. However, the performance ranking remains consistent: Tikhonov yields the lowest accuracy, TV provides moderate performance, and BM3D achieves the highest quality. Similarly, temporal variants still consistently improve over their slice-based counterparts. The greatest relative improvement is observed for BM3D, where the SSIM increases from $0.4131$ to $0.7691$, demonstrating that temporal correlation effectively mitigates noise-induced block matching errors. Temporal-BM3D always achieves the best overall performance in all methods.

\begin{table}[H]
\renewcommand{\arraystretch}{1.5}
\centering
\caption{Quantitative comparison of reconstruction from the noisy data. The best results are shown in bold.}
\begin{tabular}{|c|c|c|c|c|}
\hline
 & \multicolumn{2}{c|}{PSNR (dB)} & \multicolumn{2}{c|}{SSIM} \\
\cline{2-5}
 & Slice-Base & Temporal & Slice-Base & Temporal \\
\hline
Tikhonov & 27.2640 & 28.0137 & 0.4158 & 0.4357 \\
\hline
TV & 30.3145 & 31.2555 & 0.6062 & 0.6743 \\
\hline
BM3D & 28.1606 & \textbf{32.9085} & 0.4131 & \textbf{0.7691} \\
\hline
\end{tabular}
\label{tab:results_with_noise}
\end{table}


The optimal parameters are summarized in Table \ref{tab:N-Optimal Parameter}, showing a systematic increase in the optimal regularization strength across all priors, in contrast to the noise-free results.
In temporal models, the parameter $\gamma$ also becomes significantly higher, rising from $10^{-3}$ to $10^{-1}$ for Temporal-BM3D. These shifts indicate that stronger spatial and temporal constraints are essential to suppress noise amplification during reconstruction.

\begin{table}[H]
\renewcommand{\arraystretch}{1.3}
\centering
\caption{Optimal parameters for the noisy data.}
\begin{tabular}{|c|
>{\centering\arraybackslash}p{1.4cm}|
>{\centering\arraybackslash}p{1.4cm}|
>{\centering\arraybackslash}p{1.4cm}|}
\hline
\textbf{Parameter} & \textbf{$\lambda$} & \textbf{$\rho$} & \textbf{$\gamma$} \\
\hline
\multicolumn{4}{|c|}{\textbf{Slice-Based}} \\
\hline
Tikhonov & 0.0994 & 0.0985 & - \\
\hline
TV & 0.0479 & 0.0998 & - \\
\hline
BM3D & 17.4716 & 0.0998 & - \\
\hline
\multicolumn{4}{|c|}{\textbf{Temporal}} \\
\hline
Tikhonov & 0.0998 & 0.0939 & 0.9882 \\
\hline
TV & 0.0999 & 0.0996 & 0.0317 \\
\hline
BM3D & 18.9113 & 0.0999 & 0.6745 \\
\hline
\end{tabular}
\label{tab:N-Optimal Parameter}
\end{table}

\begin{figure*}[h]
    \centering
    \begin{tabular}{@{}c@{\hspace{25pt}}c@{\hspace{25pt}}c@{}}
        Slice-base Tikhonov & Slice-base TV & Slice-base BM3D \\[9pt]
        \includegraphics[width=0.22\textwidth]{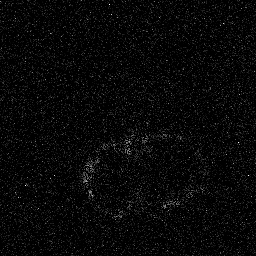} &
        \includegraphics[width=0.22\textwidth]{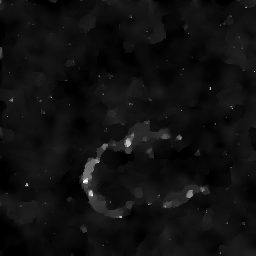} &
        \includegraphics[width=0.22\textwidth]{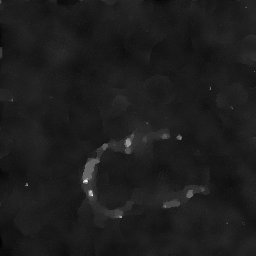} \\[9pt]   
        Temporal Tikhonov & Temporal TV & Temporal BM3D \\[9pt]
        \includegraphics[width=0.22\textwidth]{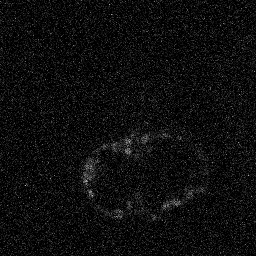} &
        \includegraphics[width=0.22\textwidth]{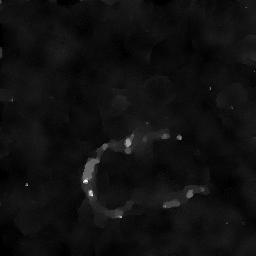} &
        \includegraphics[width=0.22\textwidth]{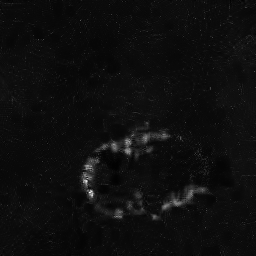}
        \\
    \end{tabular}
    \caption{Image reconstruction from noisy data, the 17th slice is plotted.}
    \label{fig:NoiseResultsImage}
\end{figure*}

Representative reconstruction results from noisy data are illustrated in Fig. \ref{fig:NoiseResultsImage}, where the same 17th slice is selected as the noise-free case. As observed, Tikhonov reconstructions suffer from severe noise contamination, with both slice-based and temporal variants failing to suppress background fluctuations and thus exhibiting low contrast and poor structural recovery. TV achieves moderate denoising, producing smoother outputs with reduced noise compared to Tikhonov, but its results are accompanied by evident edge blurring and partial loss of fine structures. BM3D demonstrates superior performance, effectively attenuating noise while preserving sharp boundaries and small details; this advantage becomes pronounced in the temporal variant, where cross-frame correlation further enhances structural consistency and background cleanliness. In general, Fig. \ref{fig:NoiseResultsImage} visually corroborates the quantitative findings: Tikhonov provides the weakest reconstructions, TV maintains an intermediate level of quality, and BM3D-particularly with temporal modeling-achieves the best balance between noise suppression and detail preservation.

\subsection{Discussion}

We further focus on two key aspects of investigation: the computational efficiency of different reconstruction methods and the reconstruction quality across varying compression ratios. The runtime analysis highlights the distinct computational costs associated with different priors, while the PSNR study provides a comprehensive view of how each method scales in accuracy under different measurement conditions.

\begin{table}[H]
\renewcommand{\arraystretch}{1.5}
\centering
\caption{Runtime Comparison.}
\begin{tabular}{|c|c|c|c|c|}
\hline
Runtime & \multicolumn{2}{c|}{Noise-Free} & \multicolumn{2}{c|}{Noisy} \\
\cline{2-5}
(\textit{sec.}) & Slice-Base & Temporal & Slice-Base & Temporal \\
\hline
Tikhonov & 0.3619 & 8.5298 & 0.6592 & 3.4593 \\
\hline
TV & 96.6382 & 128.2227 & 963.9677 & 532.5396 \\
\hline
BM3D & 852.9915 & 1722.0122 & 3751.2476 & 3555.7279 \\
\hline
\end{tabular}
\label{tab:time_comparison}
\end{table}

As shown in Table \ref{tab:time_comparison}, the three methods exhibit distinct runtime characteristics. Tikhonov achieves exceptional computational efficiency, completing slice-based reconstructions in less than a second and temporal ones within only a few seconds due to its closed-form formulation, which is up to two orders of magnitude faster than TV and hundreds of times faster than BM3D. TV, relying on iterative optimization, incurs substantially higher cost, operating on the order of minutes, but provides a favorable compromise between efficiency and reconstruction fidelity. BM3D attains the highest reconstruction accuracy at the expense of prohibitive runtime, with execution times extending to hours due to non-local block matching and collaborative filtering, and further increasing under noisy conditions where additional candidate matches are required. In the noise-free case, temporal reconstructions consistently run slower than slice-based ones, reflecting the added cost of exploiting temporal correlations, whereas under noisy conditions, this trend becomes less pronounced due to noise-induced convergence effects. These results demonstrate the fundamental trade-off between computational efficiency and reconstruction fidelity in practice.

\begin{figure}
    \centering
    \includegraphics[width=0.9\linewidth]{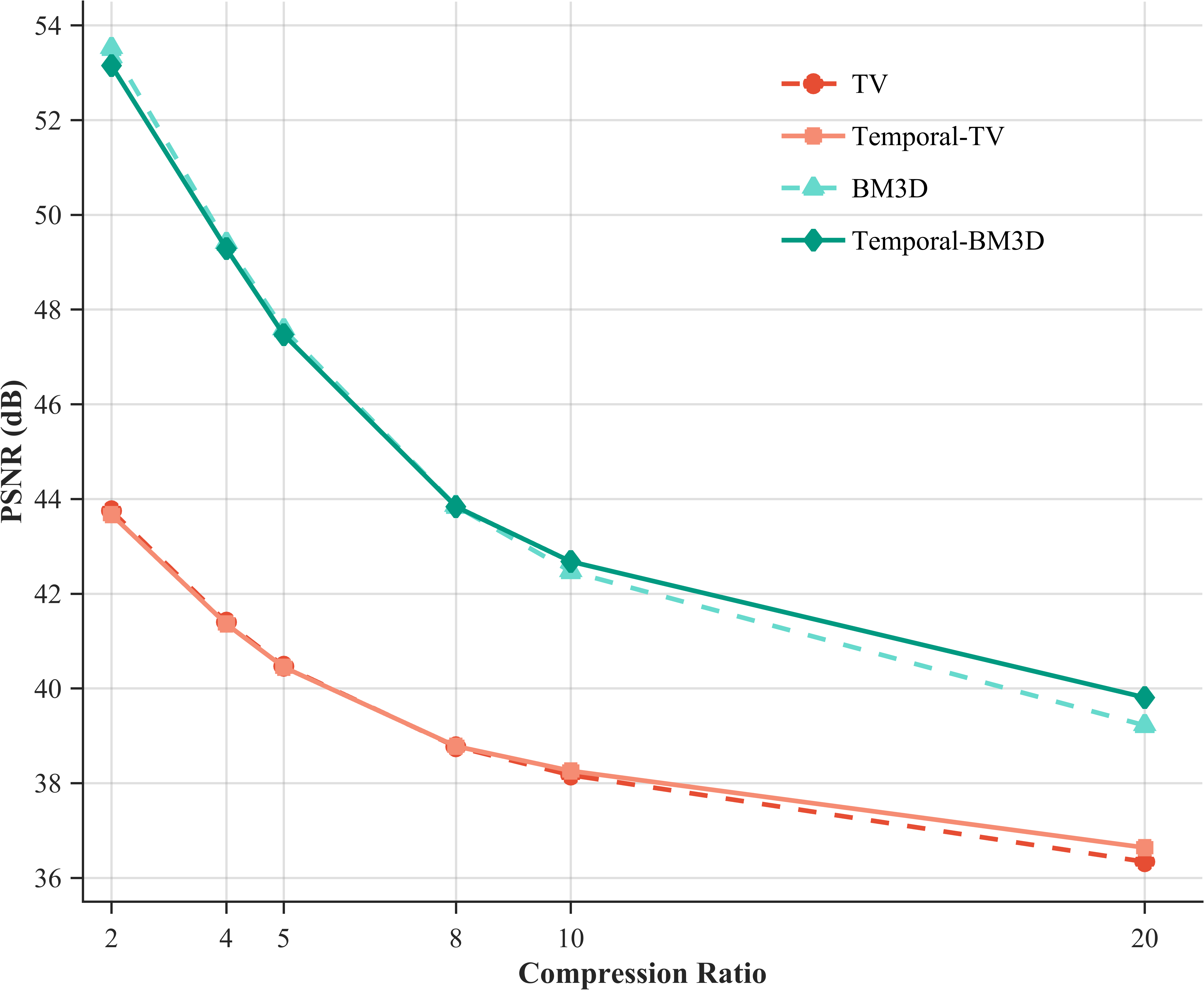}
    \caption{PSNR values for noise-free reconstructions at different compression ratios.}
    \label{fig:psnr-under-different-CR}
\end{figure}

Last but not least, we investigate the performance of four reconstruction methods (TV, Temporal-TV, BM3D, and Temporal-BM3D) under different compression ratios. As shown in Fig. \ref{fig:psnr-under-different-CR}, a smaller compression ratio corresponds to more measurements, enabling the reconstruction process to draw upon a more complete set of observations, which consistently yields higher PSNR in all four methods and highlights the advantage of achieving high-precision reconstruction under low compression conditions. Within the same type of prior, the Temporal variants demonstrate significantly better performance than their slice-based counterparts, especially at high compression ratios (CR = 10, 20), maintaining higher PSNR even under severely limited measurements. This indicates that cross-frame information fusion is particularly effective in improving reconstruction quality under extreme compression. Among all tested methods, BM3D and Temporal-BM3D outperform TV and Temporal-TV across the entire range of compression ratios, with Temporal-BM3D showing the most pronounced advantage at high compression, underscoring the strong potential of combining advanced priors with temporal modeling in compressive sensing reconstruction.

\section{Conclusion and future work}
We present a flexible PnP-ADMM reconstruction framework for CS-LSM that accommodates different image priors by plugging multiple denoisers, including Tikhonov, TV, and BM3D. One key innovation is the introduction of a temporal regularizer that exploits inter-frame correlations. This innovation yields tangible benefits, as evidenced by our results: the temporal modeling consistently improves over slice-based reconstructions. The experiments also reveal a trade-off among priors: BM3D offers superior fidelity, TV balances quality and cost, while Tikhonov achieves the fastest computation. The framework is readily extensible to learned priors and adaptive temporal weighting, providing a practical basis for high-speed, low-dose volumetric microscopy, particularly suited for dynamic biological processes such as cardiac imaging. Although this work reconstructs single 3D volumes at a fixed time point, future extensions will target simultaneous reconstruction across multiple time points of the cardiac cycle. This would enable true 4D volumetric reconstruction, capturing both spatial and temporal dynamics of the beating heart within the same framework.

\section*{Acknowledgment}
YG acknowledges support from the Kennedy Scholars Program, Department of Mathematics, University of North Carolina at Chapel Hill, through a \$2,000 research stipend. YD was partially supported by NIH R01HL162635 and NSF 2503230. YL was partially supported by NSF CAREER  DMS-2414705.

\bibliographystyle{abbrv}
\bibliography{ref.bib}

\end{document}